\documentclass{article}

\usepackage{spconf}
\usepackage{times}
\usepackage{epsfig}
\usepackage{graphicx}
\usepackage{amsmath}
\usepackage{amssymb}
\usepackage{algorithm}
\usepackage{algpseudocode}
\usepackage{epstopdf}
\usepackage{epsfig}
\usepackage{subcaption}
\usepackage{microtype}
\DeclareMathOperator*{\argmin}{arg\,min}

\title{Incorporating Scalability in Unsupervised Spatio-Temporal Feature Learning}
\name{Sujoy Paul, Sourya Roy and Amit K. Roy-Chowdhury}
\address{Dept. of Electrical and Computer Engineering, University of California, Riverside, CA 92521}

\begin{document}
%\ninept
%
\maketitle
\begin{abstract}
Deep neural networks are efficient learning machines which leverage upon a large amount of manually labeled data for learning discriminative features. However, acquiring substantial amount of supervised data, especially for videos can be a tedious job across various computer vision tasks. This necessitates learning of visual features from videos in an unsupervised setting. In this paper, we propose a computationally simple, yet effective, framework to learn spatio-temporal feature embedding from unlabeled videos. We train a Convolutional 3D Siamese network using positive and negative pairs mined from videos under certain probabilistic assumptions. Experimental results on three datasets demonstrate that our proposed framework is able to learn weights which can be used for same as well as cross dataset and tasks. {\let\thefootnote\relax\footnote{{International Conference on Acoustics, Speech, and Signal Processing (ICASSP), 2018}}}
\end{abstract}

\section{Introduction}

Large labeled datasets and computational power can be attributed as the main reason behind recent successes of Deep Neural Networks in various computer vision tasks \cite{krizhevsky2012imagenet,simonyan2014two,karpathy2015deep,he2016deep,tu2017speech,huang2018body}. 
%However, acquisition of large number of labeled samples requires intensive manual labor, which scales up rapidly for tasks involving videos. Activity recognition in streaming surveillance videos is an example, where continuous manual labeling is an infeasible task. Moreover, a huge number of unlabeled videos are available today in the internet, which can be a diverse and valuable source of visual information. 
Unsupervised learning \cite{bengio2013representation} of visual features from huge amount of unlabeled videos available today, using deep networks can be a potential solution to the data hungriness of supervised algorithms.
Autoencoders, Restricted Boltzmann Machines (RBM) and the likes \cite{bengio2007greedy,huang2007unsupervised,tramel2017deterministic} trained in a greedy layer-wise fashion have been one of the popular methods for learning visual features from images in an unsupervised manner. However, such approaches fail to discover higher level structures from the data, necessary in recognition tasks. 
%Sermanet et al. \cite{sermanet2013pedestrian} proposed a convolutional sparse coding based approach, which detects mostly semi-mid-level visual features. 

Recent works in unsupervised feature learning from image data take a slightly different path. Most of the approaches belonging to this category first define a semantically challenging task such that its training instances can be directly extracted from the unlabeled data. For example, Pathak et al. \cite{pathak2016learning} used motion information to learn visual representation of objects via a segmentation approach. Patch level puzzle solving in images has also been explored to learn visual features in \cite{doersch2015unsupervised,noroozi2016unsupervised}. Objects tracked over time played a key role in providing the supervisory signal for visual feature learning in Wang and Gupta's work \cite{wang2015unsupervised}. Ego-motion guided unsupervised learning has been explored in \cite{agrawal2015learning,jayaraman2015learning}. Li et al. \cite{li2016unsupervised} proposed an image similarity based method using low level features to learn visual features of objects in an unsupervised setting. 
\begin{figure}[t]
	\centering
	%\captionsetup{font=small}
	\includegraphics[scale=0.28]{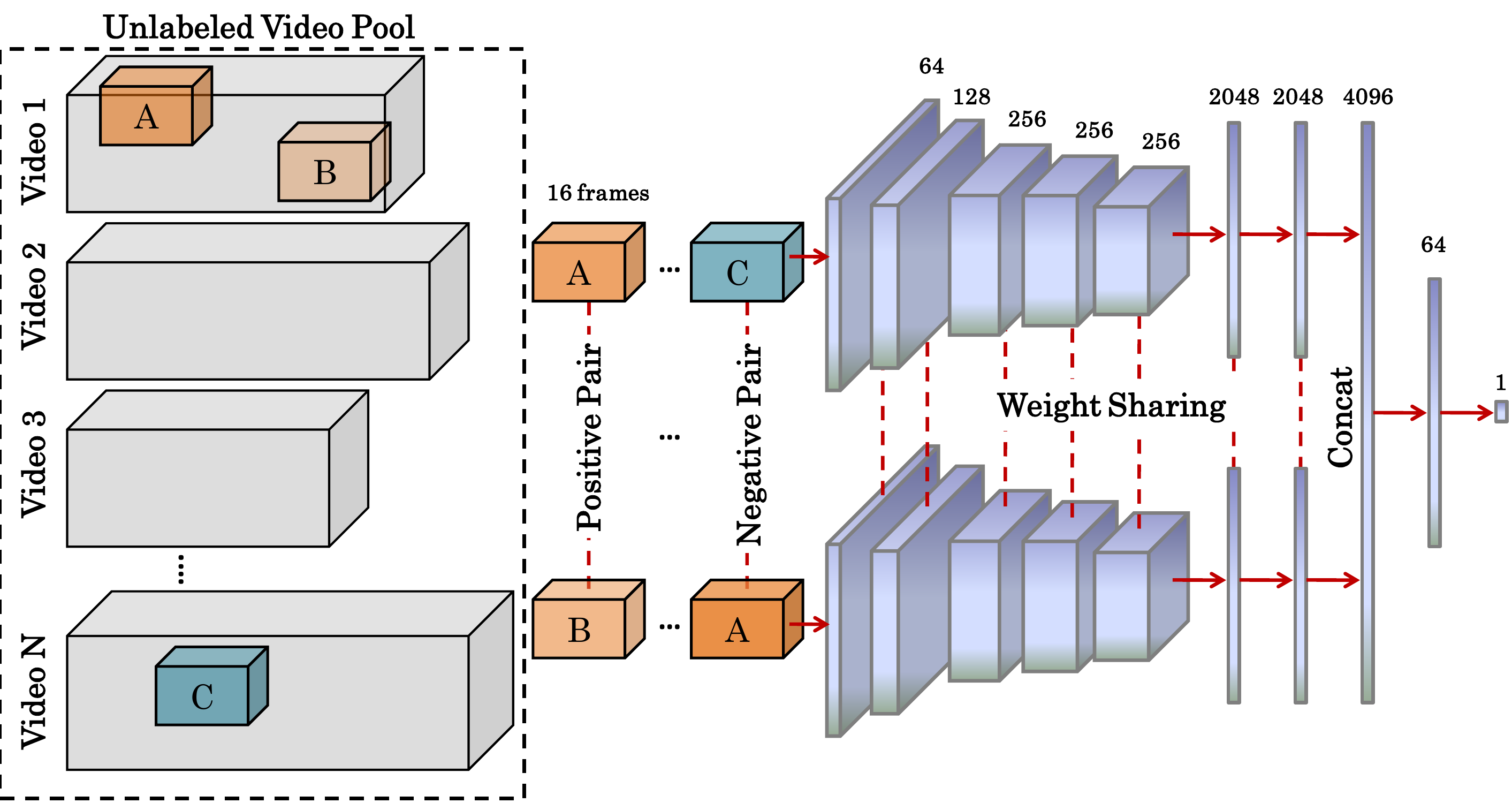}
	\caption{This figure presents our proposed framework for unsupervised feature learning. A,B,C are three example  spatio-temporal volumes. As the sub-volumes within a certain space-time boundary share similar semantic concepts, but possess different appearance and motion content, we select them as positive pairs. s-t volumes belonging to different videos constitute a negative pair (details in Section \ref{method}). }
	\label{ProposedFramework}
	\vspace{-6mm}
\end{figure}
Unlike images, unsupervised spatio-temporal representation learning from videos has not been thoroughly studied in literature. Reconstruction and prediction has been used to learn features from videos using Long Short Term Memory (LSTM) networks in \cite{srivastava2015unsupervised}. Misra et al. \cite{misra2016shuffle} exploited temporal ordering of frames in a video to learn features. It may be noted that in contrast to the previously mentioned approaches which are mainly applicable for learning only spatial features, the task of learning spatio-temporal features from videos in an unsupervised setting is significantly more challenging. The primary reason is that it is very difficult to define a tractable task for videos in the first place compared to images. %Moreover, the inherent higher dimensional nature of videos makes the unsupervised feature learning task difficult as the methods used for learning spatial representations may not be scalable to learn temporal representations in videos. 

Visual continuity is prevalent in natural %scenes. This property also prevails in natural 
videos where semantic correlation between spatio-temporally associated volumes exist \cite{dong1997spatiotemporal}. In this work, we build our hypothesis around this notion of spatio-temporal (s-t) correlatedness and argue that the likelihood of sharing higher semantic information between s-t related volumes is more compared to volumes from other videos. We structure our proposed framework based on this hypothesis to learn discriminative appearance and motion features in an entirely unsupervised manner using the 3D convolutional networks \cite{tran2015learning}. %An overview of the proposed method is discussed next.
%\noindent
%\textbf{Framework Overview.} In this paper, we propose an unsupervised learning approach to embed spatio-temporal (s-t) features. Fig. \ref{ProposedFramework} presents a pictorial representation of the proposed method. We construct our framework using a Siamese network \cite{chopra2005learning} which possesses the capability to learn data representations from pair-wise labeling instead of explicit class-wise information. We mine informative positive and negative training pairs in an unsupervised manner. Intra-video s-t volumes form the source of the positive pairs we use to train the Siamese network. On the other hand, using information extracted from across videos served as the backbone to our negative pair mining strategy. Our Siamese system incorporates weight shared 3D convolutional networks \cite{tran2015learning}, which enables our framework to learn both spatial and temporal features. Our framework is flexible from an application perspective as it can be used for unsupervised visual feature learning as well as a pre-training unit coupled with supervised tasks.
Our key contributions are below - \\
$\bullet$ We introduce a novel unsupervised feature learning framework by exploiting spatio-temporal relationships between intra and inter video segments. \\
$\bullet$ We propose an efficient strategy to mine positive and negative semantic pairs whose scalable nature makes our framework usable to ubiquitously present large unlabeled datasets. \\
$\bullet$ Finally, our proposed method can be integrated as a pre-training module with several supervised learning tasks.
\vspace{-4mm}
\section{Methodology}
\vspace{-3mm}
\label{method}
Our goal is to learn discriminative feature embedding of spatio-temporal volumes from unlabeled videos. In order to accomplish the task, we train a Siamese network based on the 3D CNN, which involves mining of positive and negative samples that can act as a supervisory data to learn semantically discriminative features. We device a simple pair mining strategy which is easy to implement and scalable to large datasets. %In order to provide a background for our proposed method, we first discuss supervised feature learning in its most general form and its extension to similarity based learning. 

\textbf{Siamese Network Training.} Let us consider that we have a set of $N$ labeled triplets $\{(\boldsymbol{x}_i^1,\boldsymbol{x}_i^2, y_i)\}_{i=1}^N$ where $\boldsymbol{x}^1_i, \boldsymbol{x}^2_i \in \mathbb{R}^m$ and $y_i \in \{+1, -1\}$. In a Siamese network \cite{hadsell2006dimensionality}, generally the same function $\mathcal{T}$ is used to project $\boldsymbol{x}_i^1,\boldsymbol{x}_i^2$ to obtain a lower dimensional representation $\boldsymbol{f}^{j}_i=\mathcal{T}(\boldsymbol{x}^j_i; \boldsymbol{W}_1)$, $j \in {1,2}$. They are converted to the desired output using another function $\mathcal{G}$. These outputs are compared with the ground-truth annotations to compute the loss $\mathcal{L}$, which needs to be minimized in order to learn the weights of the network for the particular task in hand. The optimal weights can be represented as,
\vspace{-4mm}
\begin{equation}
\boldsymbol{W}_1^*, \boldsymbol{W}_2^* = \argmin_{\boldsymbol{W}_1, \boldsymbol{W}_2} \sum_{i=1}^N \mathcal{L}(\mathcal{G}(\boldsymbol{f}^1_i, \boldsymbol{f}^2_i; \boldsymbol{W}_2), y_i)
\label{siameseloss}
\vspace{-1mm}
\end{equation}
This training strategy of Siamese networks enforces the transformation $\mathcal{T}$ to semantically group the training instances in the feature space, analogous to the perceptual grouping ability of human cognitive system. Although the process of obtaining binary labels demands lesser human effort compared to obtaining individual class labels, acquiring pair-wise labels still requires a lot of manual labeling effort. In our approach, we mine the positive and negative training pairs in an unsupervised manner as explained below. 

\textbf{Unsupervised Siamese Pair Mining.} Consider a set of $N$ unlabeled videos $\{\boldsymbol{V}_i\}_{i=1}^N$, such that $\boldsymbol{V}_i\{x,y,t,x_0,y_0,t_0\}$ denote the spatio-temporal volume of shape $(x_0, y_0, t_0)$ at position $(x,y,t)$ for the $i^{th}$ video.

\begin{figure}
	\centering
	%\captionsetup{font=small}
	\begin{subfigure}{0.20\textwidth}
		\centering
		\includegraphics[scale=0.32]{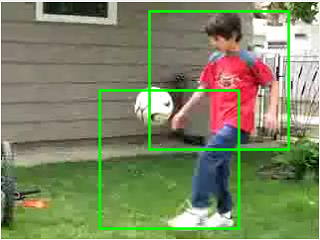}
		\caption{}
		\label{motivationRGB}
	\end{subfigure}
	\begin{subfigure}{0.20\textwidth}
		\centering
		\includegraphics[scale=0.32]{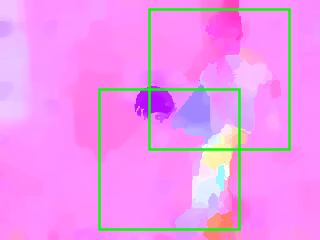}
		\caption{}
		\label{motivationFlow}
	\end{subfigure}
	\vspace{-3mm}
	\caption{This figure illustrates the motivational concept behind our positive pair mining strategy in 2D. (a) and (b) are RGB and optical flow frame of a video. The bounding boxes denote a sample positive pair. As we can observe, the appearance and motion content of the two pairs differ, but they belong to the same semantic category of \textit{Soccer Juggling}. }
	\label{motivationpositive}
	\vspace{-5mm}
\end{figure}

\textit{Positive Pair Mining.} 
In the context of training a Siamese network, positive pair can be defined as a tuple consisting of two semantically similar instances. Generally, in natural videos, distinct s-t volumes within a certain boundary contain different appearance and motion features. To elaborate, arrangements of objects across different spatial segments may vary. Similarly, separate temporal portions capture different motion structure. Despite containing different s-t patterns, such volumes represents similar semantic entities. Fig. \ref{motivationpositive} demonstrates this idea.

Although, two separate s-t volumes within a certain boundary from the same video can be used as positive pairs, the network may not be able to generalize well for videos with variations. To deal with this issue, we expand our pool of positive pairs by applying the following transformations- \textbf{1.} Color transform $\mathcal{F}_1$ in the HSV domain involving three parameters (discussed in Section \ref{experiments}), \textbf{2.} A non-parametric transformation $\mathcal{F}_2$ defined as $XA$, where $A$ is an anti-diagonal matrix containing only $1$s and $X$ is an image. This operation flips the image in the horizontal direction.
It may be noted that the same transformation is carried out in all the images of the spatio-temporal volume. Finally, the poitive pairs we use may be defined as \vspace{-2mm}
\[\{\boldsymbol{V}_i\{x^1,y^1,t^1,x_0,y_0,t_0\}, \mathcal{F}(\boldsymbol{V}_i\{x^2,y^2,t^2,x_0,y_0,t_0\})\}\]
where $\mathcal{F}$ is chosen probabilistically from the set of transformations $\{\mathcal{I}, \mathcal{F}_1, \mathcal{F}_2, \mathcal{F}_1 \circ \mathcal{F}_2\}$ and $\mathcal{I}$ is the identity transformation. The probabilities associated with selection of each transformation can be found in Algorithm \ref{algo:framework}. 

\textit{Negative Pair Mining.} Our negative pair mining strategy is based on the following idea. Consider that the unlabeled video pool came from $m$ distinct distributions representing semantic concepts and $n_k$, $k \in \{1, \dots, m\}$ be the number of instances belonging to the $k^{th}$ distribution. The maximum probability that a pair of s-t volumes extracted from two videos randomly from the entire dataset, belong to the same distribution is \vspace{-2mm} 
\[p \leq \Big(\frac{\max_{k} n_k}{\sum_{k=1}^m n_k}\Big)^2m\]
Assuming that $\max_{k} n_k \texttt{<<} \sum_{k=1}^m n_k$, which may be the case in natural unlabeled video pool, $p_{max} \rightarrow 0$. With this assumption, the negative pairs may be defined as, \vspace{-2.5mm}
\[\{\boldsymbol{V}_i\{x^1,y^1,t^1,x_0,y_0,t_0\}, \mathcal{F}(\boldsymbol{V}_{j \ne i}\{x^2,y^2,t^2,x_0,y_0,t_0\})\}\]

\textbf{Learning Spatio-Temporal Features.} Learning spatio-temporal features for videos is important for several recognition tasks in computer vision. Convolutional 3D (C3D) \cite{tran2015learning} network have been successful in learning both motion and appearance features. We use the smaller C3D network defined in their paper as a transformation $\mathcal{T}$ in Eqn. \ref{siameseloss}. The output of this transformation is feature vectors $\boldsymbol{f}^1, \boldsymbol{f}^2 \in \mathbb{R}^{2048}$ for the two input s-t volumes of the Siamese network. We concatenate the two feature vectors to obtain a single feature vector $\in \mathbb{R}^{4096}$. In order to learn the transformation $\mathcal{G}:\mathbb{R}^{4096} \rightarrow \mathbb{R}$ in Eqn. \ref{siameseloss}, we use two fully connected (fc) layers such that output of first fc layer $\in \mathbb{R}^{64}$. Finally, we minimize the hinge loss \cite{rosasco2004loss} for binary classification, which may be presented as,
\vspace{-3mm}
\begin{equation}
\mathcal{L} = \sum_{i=1}^B C_i\max(0,1-t_iy_i) + \lambda ||\boldsymbol{W}||_F
\label{hingeloss}
\vspace{-2mm}
\end{equation}
where $t_i$ is the scalar output of the siamese network for the $i^{th}$ pair in the batch and $\boldsymbol{W}$ represents all the weights of the network. The second part of the equation is the regularization term and we set $\lambda=0.0005$ in our experiments. $B$ is the mini-batch size of Stochastic Gradient Descent.  $C_i$ is $\frac{1}{N_p}$ and $\frac{1}{N_n}$ respectively for number of positive and negative samples in the mini-batch.  $t$ is the final scalar output of the Siamese network. We use dropout  of $0.5$ in the $fc$ layers, except the final output layer. 

\begin{figure*}[t]
	\centering
	%\captionsetup{font=small}
	\begin{subfigure}{0.11\textwidth}
		\includegraphics[height=15mm,width=18mm]{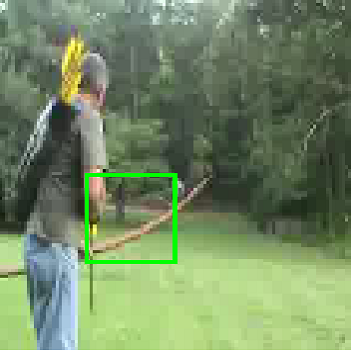}
		\label{}
	\end{subfigure}
	\begin{subfigure}{0.11\textwidth}
		\includegraphics[height=15mm,width=18mm]{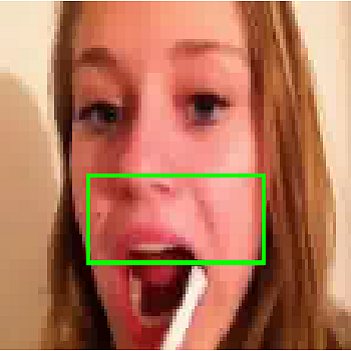}
		\label{}
	\end{subfigure}
	\begin{subfigure}{0.11\textwidth}
		\includegraphics[height=15mm,width=18mm]{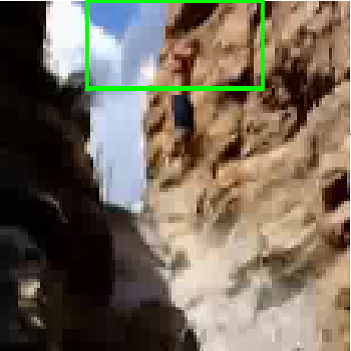}
		\label{}
	\end{subfigure}			
	\begin{subfigure}{0.11\textwidth}
		\includegraphics[height=15mm,width=18mm]{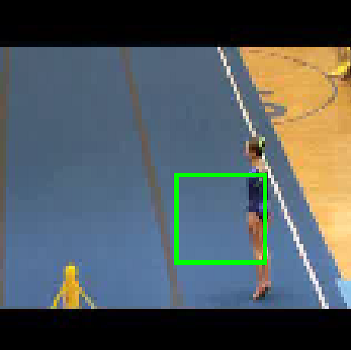}
		\label{}
	\end{subfigure}		
	\begin{subfigure}{0.11\textwidth}
		\includegraphics[height=15mm,width=18mm]{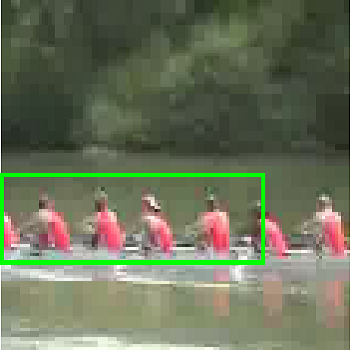}
		\label{}
	\end{subfigure}
	\begin{subfigure}{0.11\textwidth}
		\includegraphics[height=15mm,width=18mm]{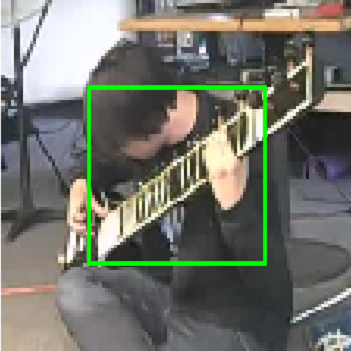}
		\label{}
	\end{subfigure}	
	\begin{subfigure}{0.11\textwidth}
		\includegraphics[height=15mm,width=18mm]{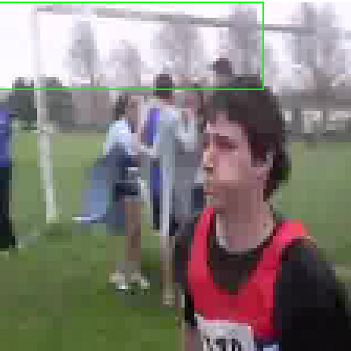}
		\label{}
	\end{subfigure}		
	\begin{subfigure}{0.11\textwidth}
		\includegraphics[height=15mm,width=18mm]{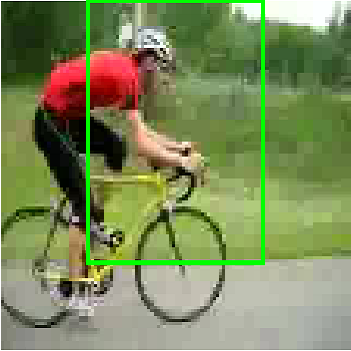}
		\label{}
	\end{subfigure}	
	\vspace{-3mm}
	\caption{This figure presents the top response region of pool5 of our network learned from unlabeled data. }
	\label{bbox}
\end{figure*}

\begin{table*}
	%\captionsetup{font=small}
	\small
	\centering
	\caption{Activity recognition accuracy on UCF101 and HMDB51. The column with C3D indicate results when the network is trained using randomly initialized weights. 'Ours' for HMDB51 mean UCF101 Unsupervised + HMDB51 Supervised, and that for UCF101 mean UCF101 Unsupervised + UCF101 Supervised. }
	\vspace{-2mm}
	\begin{tabular}{|| c | c | c | c | c | c | c | c |c ||}
		\hline
		{Algorithms $\rightarrow$} & {Chance} & {C3D \cite{tran2015learning}} & {STIP \cite{soomro2012ucf101}} & {TCH \cite{mobahi2009deep}} & {IM \cite{hadsell2006dimensionality}} & {OP \cite{wang2015unsupervised}} & {S\&L \cite{misra2016shuffle}} & {Ours} \\
		\hline \hline
		{HMDB51} & {1.9} & {19.2} & {\underline{20.0}} & {15.9} & {16.3} & {15.6} & {18.1} & {\textbf{20.6}} \\
		\hline
		{UCF101} & {01.0} & {44.1} & {43.9} & {45.4} & {45.7} & {40.7} & {\textbf{50.2}} & {\underline{48.5}} \\
		\hline \hline
	\end{tabular}
	\label{activity}
	\vspace{-3mm}
\end{table*}

\begin{algorithm}[h]
	\small
	\caption{Online Pair Mine Algorithm}
	\label{algo:framework} 
	\begin{algorithmic}
		\State {\bf Input:} 1. Unlabeled video dataset $\{\boldsymbol{V}_i\}_{i=1}^N$, \\ \hspace{10.5mm}2. Positve Pair Ratio ($p$)
		\State {\bf Output:} Training data batch $\{\boldsymbol{S}^1_i,\boldsymbol{S}^2_i,y_i\}_{i=1}^B$ 
		\State \textbf{1.} $N_p = \text{round}(pB)$, $N_n = B - N_p$, $i \leftarrow 1$
		\While{$i \leq N_p$}
		\State $m \sim \mathcal{U}[1,N]$
		\State $\boldsymbol{S}^1_i,\ \boldsymbol{S}^2_i \leftarrow $ $\text{STVolume}(\boldsymbol{V}_m), \ \text{STVolume}(\boldsymbol{V}_m)$
		%\State \textbf{5.} $\boldsymbol{S}^2_i \leftarrow $ $\text{STVolume}(\boldsymbol{V}_m)$
		\If{$\boldsymbol{S}^1_i \cap \boldsymbol{S}^2_i \neq \Phi$}
		\State Goto Step \textbf{5}
		\EndIf
		\State $y_i \leftarrow +1$, $i \leftarrow i + 1$
		\EndWhile
		\While{$i \leq N_n$}
		\State $m \sim \mathcal{U}[1,N]$, $n \sim \mathcal{U}\{[1,N]-m\}$
		\State $\boldsymbol{S}^1_i,\ \boldsymbol{S}^2_i \leftarrow $ $\text{STVolume}(\boldsymbol{V}_m),\ \text{STVolume}(\boldsymbol{V}_n)$
		%\State \textbf{5.} $\boldsymbol{S}^2_i \leftarrow $ $\text{STVolume}(\boldsymbol{V}_n)$
		\If{$r_1 \sim \mathcal{N}(0,1) > 0$}
		\State $\boldsymbol{S}^2_i \leftarrow \mathcal{F}_1(\boldsymbol{S}^2_i)$
		\EndIf
		\If{$r_2 \sim \mathcal{N}(0,1) > 0$}
		\State $\boldsymbol{S}^2_i \leftarrow \mathcal{F}_2(\boldsymbol{S}^2_i)$
		\EndIf
		\State $y_i \leftarrow -1$, $i \leftarrow i + 1$
		\EndWhile	
	\end{algorithmic} 
	\textbf{Note}: $\text{STVolume}(\boldsymbol{V})$ is a function which randomly extracts a spatio-temporal volume from the video $\boldsymbol{V}$. 
	\label{algorithm}
\end{algorithm}
\vspace*{-6mm}

\section{Experiments and Results}
\vspace*{-1mm}
\label{experiments}
In this section, we present results and analysis of the proposed unsupervised feature learning algorithm. We mainly focus on the activity recognition and video similarity classification.

\textbf{Unsupervised Learning.} We use the UCF101 \cite{soomro2012ucf101} dataset for training our C3D Siamese network in an unsupervised manner. We follow Algorithm \ref{ProposedFramework} to mine the pairs required for training our C3D Siamese network. To optimize the loss function, we use Adam Optimizer \cite{kingma2014adam} in a Stochastic Gradient Descent setting with mini-batch size of 10 and 30-70\% split in positive and negative samples respectively. %We use learning rate of $1e-4$. with step decay by a factor of $2$ applied every 20k iterations. We train the network for 50k iterations. We initialize our network randomly with xavier initialization method \cite{glorot2010understanding}. We implement all our algorithms in TensorFlow \cite{abadi2016tensorflow} and use a Tesla K80 GPU to train the network. \footnote{Codes will be made available online.} 

\textit{Color Transformations.} The color transformation $\mathcal{F}_1$ mentioned in Section \ref{method} are as follows. \textbf{1.} Adding a random number $\in (-0.1, 0.1)$ to all the pixels of an image, \textbf{2.} Taking element-wise exponent of the Saturation and Value component with a random number $\in (0.5, 2)$, \textbf{3.} Scaling the Saturation and Value components by a random number $\in (0.7, 2)$. All random numbers are generated from uniform distributions.

\textit{Feature Representation Visualization.} We visualize the $pool5$ feature response of our unsupervised C3D network to identify the regions it detect as semantically relevant. We visualize only single frame in Fig. \ref{bbox}. We obtain $256$ activation maps from the pool5 layer. Then, we average over all the receptive fields corresponding to the units with maximum response from each activation map. Finally, we segment the averaged receptive fields to obtain the bounding boxes. Results depict that our network is able to identify areas of the image which involve human-object interaction and motion. 

\begin{figure*}[h!]
	\centering
	\includegraphics[scale=0.40]{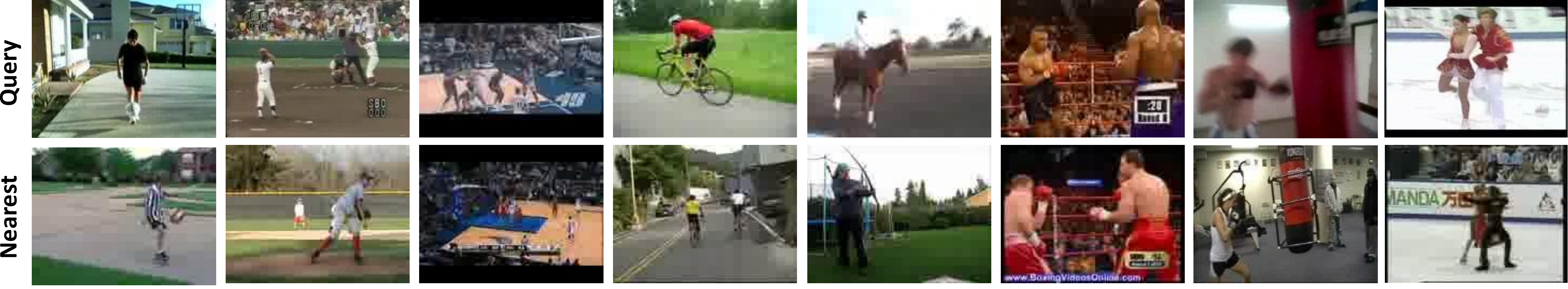}
	\caption{In this figure we present the nearest neighbors of a query video using the features learned by our unsupervised network.}
	\label{pairs}
	\vspace{-2mm}
\end{figure*}

\textit{Nearest Neighbor.} In order to understand the semantic concepts learned by our network trained in an unsupervised fashion, we retrieve the nearest neighbor of query videos. For each video of the UCF101 dataset, we extract $10$ random s-t volumes and pass it through the learned network to obtain $10$ feature vectors $\in \mathbb{R}^{2048}$ from the $fc7$ layer followed by mean pooling to obtain a single feature vector. Then, given a query video, we find its nearest neighbor in the feature space. Sample results of the nearest neighbor video retrieval task are presented in Fig. \ref{pairs}. This shows that although the positive training pairs belong to different spatio-temporal volumes of the same video of UCF101, the network is able to generalize to similar semantic concepts belonging to different videos. 
\begin{figure}[h!]
	\vspace{-4mm}
	\centering
	\includegraphics[scale=0.35]{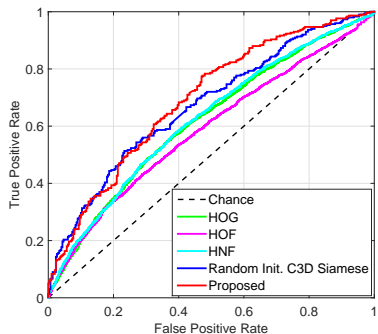}
	\caption{ROC curve on ASLAN dataset}
	\label{roc}
	\vspace{-3mm}
\end{figure}
\begin{table}
	\centering	
	\caption{Pair mining computational time for a training sample}
	\vspace{-2mm}
	\begin{tabular}{||c|c|c||} 
		\hline
		{Algorithms} & {Optical Flow} & {Ours} \\
		\hline \hline
		{Time per pair (in second)} & {$8.0\times 10^{-2}$} & {$1.8\times 10^{-5}$} \\
		\hline
	\end{tabular}
	\label{table:2}
	\vspace{-4mm}
\end{table}
\begin{table}[h]
	\vspace{-1mm}
	\footnotesize
	\centering	
	\caption{Action Similarity results on ASLAN}
	\vspace{-3mm}
	\begin{tabular}{||c|c|c|c|c|c||} 
		\hline
		{Algo.}&{C3D \cite{tran2015learning}}&{HOG \cite{kliper2012action}}&{HOF  \cite{kliper2012action}}& {HNF  \cite{kliper2012action}}&{Ours}\\
		\hline \hline
		{Acc.}&{57.3}&{56.6}&{56.8}&{58.9}&{\textbf{63.5}} \\
		\hline
		{AUC}&{67.2}&{61.6}&{58.5}&{62.1}&{\textbf{69.3}} \\
		\hline
		%{\begin{tabular}{c} 
		% UCF101 Unsup. + ASLAN Unsup.  \\ 
		% +  Finetune (Proposed)
		% \end{tabular}} & {} \\
		%\hline
	\end{tabular}
	\label{table:3}
	\vspace{-4mm}
\end{table}

\textbf{Finetuning.} A deep neural network is generally trained on a large number of labeled instances and then finetuned on a smaller task-specific dataset \cite{sharif2014cnn}. In scenarios where constrained budget limit the acquisition of labeled samples, we may have to train the network using only a small task-specific dataset, which may not yield good performance. %However, it may be advantageous to utilize large unlabeled datasets, which are generally easily available, in the learning process in order to achieve better performance. 
In this section, we demonstrate that the proposed method can be used to first learn a feature embedding in an unsupervised way from a large unlabeled dataset, and then the network weights can be finetuned on a smaller task-specific dataset to obtain better performance compared to a randomly initialized network. We also show that unsupervised feature learning using our framework on a dataset followed by finetuning with labels on the same dataset can perform better compared to supervised training from randomly initialized weights. In this section, we use the HMDB51 dataset \cite{kuehne2011hmdb} for activity recognition which is a smaller dataset than UCF101.

\textit{Results.} We use the learned weights from our unsupervised Siamese network, trained on UCF101, to finetune using the HMDB51 dataset. We also train the C3D network from randomly initialized weights using the HMDB51 dataset. We compare our results against four baseline methods for unsupervised feature learning, which are - Temporal Coherence (TCH) \cite{mobahi2009deep}, Invariant Mapping (IM) \cite{hadsell2006dimensionality}, Object Patch (OP) \cite{wang2015unsupervised} and Shuffle \& Learn (S\&L) \cite{misra2016shuffle}. The results are presented in Table \ref{activity}. Better performance of the proposed method over the randomly initialized network suggests that the proposed unsupervised learning framework can serve as a module to structure the feature space for superior supervised learning performance. We also finetune the network weights learned in an unsupervised manner using UCF101 class labels. We compare with other unsupervised baselines mentioned previously. The results are presented in Table \ref{activity}. As can be observed that, the proposed method performs better than randomly initialized C3D by a \textbf{margin of $\textbf{4.4\%}$}, which clearly indicates that our method can be used to enhance the performance of video related supervised tasks. Our method also performs better than other baselines except S\&L. However, S\&L and other works in literature involve optical flow in their pair mining strategy which adds to the computational time. On the other hand, our randomized pair mining strategy is computationally efficient and scalable to large datasets. Table \ref{table:2} presents this comparison. 

\textbf{Action Similarity Classification}
Our proposed unsupervised framework learns similarities in videos which can be used to solve the problem of similarity labeling, given a pair of videos. In this section, we explore, whether our network, learned in an unsupervised manner, can help to achieve better performance compared to randomly initialized network on this task. We use the Action Similarity Labeling (ASLAN) dataset \cite{kliper2012action} for this experiment. 

\textit{Results.} We present the results of our network on ASLAN dataset along with the result obtained after training the Siamese C3D network from randomly initialized weights in Table \ref{table:3}. We also compare the results with other baselines from existing literature \cite{kliper2012action}. The ROC curve is presented in Fig. \ref{roc}. It is evident from the experimental results that the proposed method performs better than randomly initialized Siamese C3D network which suffers from the scarcity of training data.
\vspace*{-2mm}
\section{Conclusion}
\vspace*{-2mm}
In this work we present a novel approach to learn spatio-temporal feature learning from unlabeled videos. Experimental results suggest that the embeddings learned by our framework are transferable to new datasets and can be finetuned to achieve superior performance than training with random initialization using the new dataset. Furthermore, the performance of supervised learning on a certain dataset can be improved by first using our unsupervised learning scheme on the same dataset.
\textbf{Acknowledgment.} This work was partially supported by NSF grants  IIS-1316934 and 1544969.
%Future work will consider using unsupervised training for various visual representation learning problems.
%In future, we plan to exploit the multi-model data associated with videos available in the internet. Specifically, the tags corresponding to the web-videos can serve as a rich source of self-supervisory signal for unsupervised visual representation learning. The problem of key frame extraction from unlabeled videos is also an interesting direction which can explored by building on top of our proposed framework.  

\bibliographystyle{ieee}
\bibliography{refs}

\end{document}